\documentclass[a4paper,table]{article}
\usepackage{spconf}
\usepackage{amsmath,graphicx}
\usepackage{hhline} 
\usepackage{etoolbox}
\usepackage{pgf}
\usepackage{xcolor}
\usepackage{multirow}
\usepackage{booktabs}
\usepackage{array}
\usepackage{tabularx}
\usepackage{arydshln}
\usepackage[T1]{fontenc}
\usepackage{amssymb}
\usepackage{comment}
\usepackage{fancyhdr}

\newcolumntype{?}{!{\vrule width 1pt}}
\DeclareMathOperator*{\argmin}{arg\,min}

\title{Audio-AdapterFusion: A Task-ID-free Approach for Efficient and Non-Destructive Multi-task Speech Recognition}
\name{Hillary Ngai, Rohan Agrawal, Neeraj Gaur, Ronny Huang, Parisa Haghani, Pedro Moreno Mengibar}
\address{Google LLC}

\begin{document}
\copyrightnotice{979-8-3503-0689-7/23/\$31.00~\copyright2023 IEEE}
\maketitle
\begin{abstract}

Adapters are an efficient, composable alternative to full fine-tuning of pre-trained models and help scale the deployment of large ASR models to many tasks. In practice, a task ID is commonly prepended to the input during inference to route to single-task adapters for the specified task. However, one major limitation of this approach is that the task ID may not be known during inference, rendering it unsuitable for most multi-task settings. To address this, we propose three novel task-ID-free methods to combine single-task adapters in multi-task ASR and investigate two learning algorithms for training. We evaluate our methods on 10 test sets from 4 diverse ASR tasks and show that our methods are non-destructive and parameter-efficient. While only updating 17\% of the model parameters, our methods can achieve an 8\% mean WER improvement relative to full fine-tuning and are on-par with task-ID adapter routing.

\end{abstract}
\section{Introduction \& Background}
\label{sec:intro}

The most commonly used method for training Automatic Speech Recognition (ASR) systems is to perform transfer learning \cite{kunzeKKKJS17}. Typically, a state-of-the-art ASR model, such as the Conformer \cite{gulati2020conformer}, is pre-trained on a source task with semi-supervised learning \cite{zhang2020pushing, lin2022survey}. The pre-trained model is then adapted to the target task by fine-tuning \emph{all of its weights} on this \emph{single task} \cite{guo2021toolkit, grouchy2020transformer}. While this approach can achieve impressive results on one task, fine-tuning the entire model per task is computationally expensive and does not scale well when there are many tasks \cite{thomas2022efficient}.

In prior works, researchers have attempted to solve this problem using parameter-sharing methods, such as sequential fine-tuning and multi-task learning (MTL) \cite{kim2017asrmultitask, tang2021general}. Sequential fine-tuning involves pre-training the model on the source task and fine-tuning the entire model on the target tasks one after the other. At each fine-tuning step, the model is initialized with the parameters learned from the previous task \cite{phang2018sentence, ngai2021transformer}. However, previous research has shown that this approach exhibits poor performance beyond two sequential tasks due to catastrophic forgetting \cite{phang2018sentence, ramasesh2022effect}. Catastrophic forgetting, also known as catastrophic interference, is a phenomenon that occurs when a model abruptly loses information of the previous learned task(s) after training on a new task \cite{kirkpatrick2017overcoming}.

\begin{figure}[htb]
\centering
\centerline{\includegraphics[width=4.5cm]{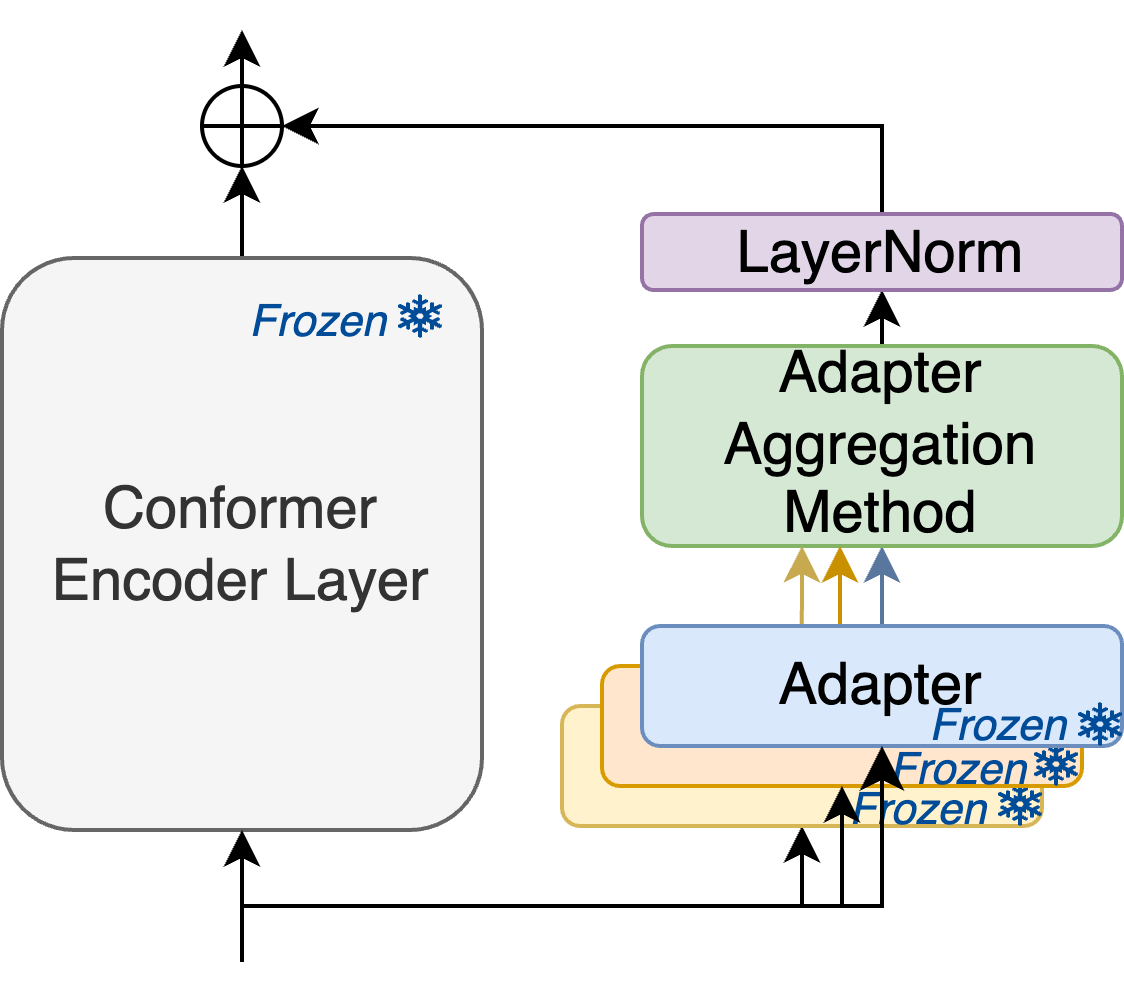}}
\caption{Adapter aggregation architecture parallel to a frozen conformer encoder layer. The adapter aggregation method takes as input the representations of multiple parallel adapters trained on different tasks and learns a representation that combines useful information from each adapter.}

\label{fig:adapterfusion-conformer}
\end{figure}

Unlike sequential fine-tuning, MTL trains a single model on all of the tasks simultaneously by combining the task objectives \cite{ruder2017overview, gaur2021mixture}. The most common method to combine task objectives in MTL is to take the mean loss over all of the tasks \cite{gong2019comparison}. By leveraging shared parameters that capture common and useful information across tasks, MTL aims to improve the performance of multiple related tasks. However, MTL requires simultaneous access to all of the tasks during training. This means that when new tasks are added, all of the shared weights will need to be retrained \cite{zhang2021e2e, ruder2017overview}. Furthermore, directly optimizing the average loss could lead to degradation in performance of certain tasks due to conflicting gradients from different tasks. Gradients from different tasks can exhibit different magnitudes, with the largest gradient dominating the model update. These gradients may also point in opposing directions, causing the model update to diverge from the local optima of certain tasks \cite{liu2021conflict, ngai2022doctor}.

Recently, residual adapters \cite{houlsby2019parameter} have emerged as an efficient, composable alternative to full fine-tuning of pre-trained models. Adapters were initially proposed for language modeling but have also shown success in ASR \cite{thomas2022efficient, hou2021exploiting, vander2023using, li23fa_interspeech}. Due to its parameter-efficiency and modularity, adapters are particularly useful to scale the training and serving of large ASR models to many tasks \cite{thomas2022efficient}. Instead of fine-tuning the entire ASR model for each task, a single-task adapter \textemdash consisting of a relatively small number of randomly-initialized parameters \textemdash is introduced at every Conformer encoder layer for each task. While freezing the weights of the shared pre-trained model, single-task adapters are trained separately for multiple tasks. Despite only training a few additional parameters per task, adapters have been shown to perform on-par with full fine-tuning \cite{thomas2022efficient, hou2021exploiting, vander2023using}. In production settings, a task ID is commonly prepended to the input during inference to route to only the adapters trained on that specified task \cite{houlsby2019parameter, thomas2022efficient}. However, \emph{one major limitation of this approach is that the task ID is typically unknown during inference time}. Moreover, this approach restricts the knowledge sharing between adapters trained on different tasks, thereby impeding the model's potential for enhancing generalizability. Thus, using a task ID is not practical for most multi-task settings.

Recently, AdapterFusion \cite{pfeiffer2020adapterfusion} was proposed to address these issues in language modeling. AdapterFusion is a task-ID-free method that leverages knowledge from different single-task adapters to solve multiple tasks, without suffering from the same problems as sequential fine-tuning and MTL. There are two stages in AdapterFusion \textemdash the knowledge extraction stage and the knowledge composition stage. In the knowledge extraction stage, single-task adapters are trained separately on multiple tasks. In the knowledge composition stage, adapters trained on various tasks are combined to share information across tasks.

We propose Audio-AdapterFusion (A-AF), a novel adaptation of AdapterFusion to multi-task ASR. A-AF combines parallel adapters trained on different tasks in an efficient and non-destructive manner to solve multiple ASR tasks. Our task-ID-free method outperforms full fine-tuning and is on-par with using a task ID to route to adapters trained on the specified task.

\subsection{Contributions}
\begin{figure}[htb]
\centering
\centerline{\includegraphics[width=5.5cm]{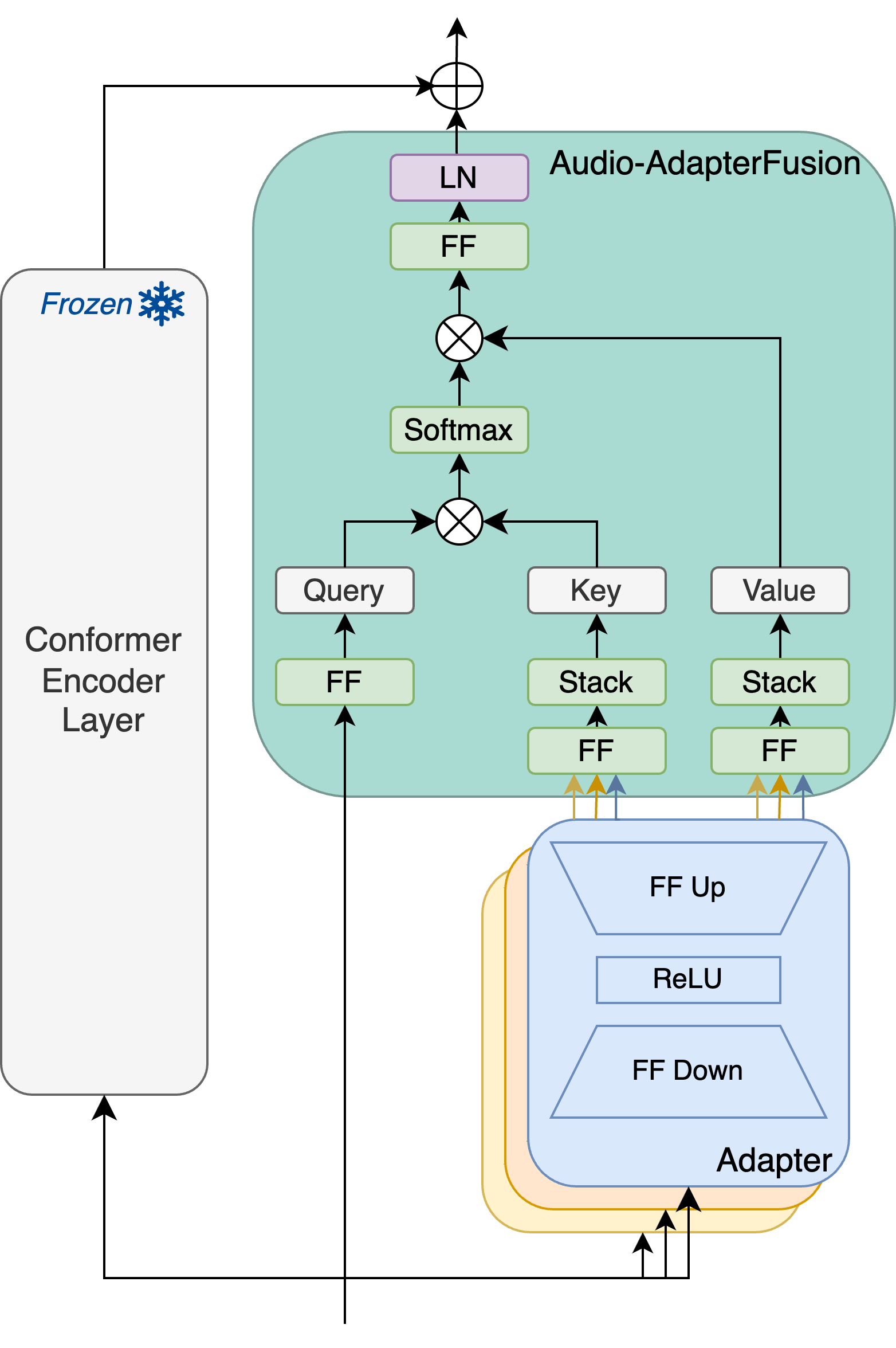}}
\caption{Our Audio-AdapterFusion architecture. This includes the query, key, and value. The query is the projected input of the frozen Conformer encoder layer. Both key and value are the projected and stacked outputs of the parallel adapters. The dot product of the query and key is passed through a softmax function to learn to weight each feature of each adapter, given an utterance. The projected weighted value is passed through a LayerNorm before being added to the Conformer output via a residual layer.}
\label{fig:adapterfusion}
\end{figure}

Our main contributions are:
(1) We present three novel methods to combine single-task adapters to solve multiple ASR tasks: \emph{Adapter Mean (Avg)}, \emph{Adapter Weighted Mean (WAvg)}, and \emph{Audio-AdapterFusion (A-AF)}. These methods do not require a task ID.
(2) Unlike AdapterFusion, our methods combine parallel adapters \textemdash instead of sequential adapters \textemdash at each Conformer encoder layer \textemdash rather than BERT layer \textemdash and add Layer Normalization (LayerNorm) \cite{ba2016layer} before the residual layer as shown in Fig. ~\ref{fig:adapterfusion}. We empirically find that LayerNorm before the residual layer improves training stability and performance of AdapterFusion.
(3) We explore two different learning algorithms to train our adapter aggregation methods: with or without updating and sharing the pre-trained adapter weights in the knowledge composition stage.
(4) We evaluate our methods on 10 test sets from a set of 4 diverse ASR tasks: closed captioning, short-form audio from call centers, long-form audio from call centers, and speech search. (4) We show that our best-performing methods significantly outperform full fine-tuning on multiple tasks \textemdash including a zero-shot learning task \textemdash and are on-par with task-ID adapter routing.
(5) Finally, we analyze the trade-off between performance and the number of trained parameters in the knowledge composition stage of our proposed methods. We provide diverse combinations of adapter aggregation methods and learning algorithms to cater to different requirements and limitations of AI practitioners.

\section{Audio-AdapterFusion}
\label{sec:audio-adapterfusion}

To date, there are no ASR solutions that use adapters for multiple tasks without the need for a task ID or without suffering from catastrophic interference. To address this, \emph{we propose Audio-AdapterFusion, a parameter-efficient, task-ID-free method for multi-task ASR in a non-destructive manner}.

\subsection{Learning Algorithm}
In the first stage of our learning algorithm, we freeze the pre-trained base model $\Theta$ and introduce $N$ single-task adapter (ST-A) \cite{houlsby2019parameter} modules at each Conformer encoder layer $l$. As shown in Eq. \eqref{eqn:single-task-adapter}, we train each adapter module $\Phi_{n}$ on their respective task $D_{n}$ with RNN Transducer (RNN-T) loss $\mathcal{L}$ \cite{graves2012sequence}. This is called the knowledge extraction stage \cite{pfeiffer2020adapterfusion}.

\begin{equation}
\label{eqn:single-task-adapter}
\Phi_{n} \leftarrow \argmin_{\Phi_{n}} \mathcal{L}(D_{n}; \Theta, \Phi_{n})
\end{equation}

In the second stage of our learning algorithm, we combine the task-specific knowledge to solve multiple tasks. Specifically, we freeze the pre-trained base model parameters $\Theta$ and all of the pre-trained single-task adapter parameters $\Phi$. We combine the $N$ task adapters at each Conformer encoder layer $l$ to solve multiple tasks $D$. Depending on the adapter aggregation method, described in section \ref{sec:adapter-aggregation}, we may introduce fusion parameters $\Psi$ at each Conformer encoder layer to learn to combine the outputs of the single-task adapters. We train the fusion parameters $\Psi$ on multiple tasks with RNN-T loss $\mathcal{L}$. This is called the knowledge composition stage \cite{pfeiffer2020adapterfusion}.

\begin{equation}
\label{eqn:adapter-aggregation-loss}
\Psi \leftarrow \argmin_{\Psi} \mathcal{L}(D; \Theta, \Phi_{1}, ..., \Phi_{N}, \Psi)
\end{equation}

By separating the two training stages \textemdash knowledge extraction and knowledge composition \textemdash we address the issues of catastrophic interference between different tasks for multi-task ASR.

\subsection{Adapter Aggregation Methods}
\label{sec:adapter-aggregation}

We propose three different methods of aggregating the $N$ adapter outputs at each Conformer encoder layer: \emph{Adapter Mean (Avg)}, \emph{Adapter Weighted Mean (WAvg)}, and \emph{Audio-AdapterFusion (A-AF)}. We define below each adapter aggregation method where $h_{l, t, i, n}$ is the output of the $n^{th}$ task adapter at layer $l \in \{1, ..., L\}$, time step $t \in \{1, ..., T\}$, and hidden dimension $i \in \{1, ..., d_{model}\}$.

\begin{enumerate}
\itemsep0em 
  \item \emph{Adapter Mean (Avg)}: the LayerNorm of the element-wise mean of the adapter outputs at Conformer encoder layer $l$. Avg requires no added parameters and therefore no additional training.
\begin{equation}
\label{eqn:a-avg-pre}
\textbf{A}_{l, t, i} =  \frac{1}{N}\sum_{n=1}^{N} h_{l, t, i, n}
\end{equation}
\begin{equation}
\label{eqn:a-avg}
\textbf{Avg}_{l} =  LayerNorm(\textbf{A}_{l})
\end{equation}
  \item \emph{Adapter Weighted Mean (WAvg)}: the LayerNorm of the weighted element-wise mean of the adapter outputs at Conformer encoder layer $l$ where the weights $w_{l, n} \forall \; l\in\{1, ..., L\}, n\in\{1, ..., N\}$ are introduced in the knowledge composition stage and trained to solve multiple tasks. 
\begin{equation}
\label{eqn:a-wavg-pre}
\textbf{A}_{l, t, i} =  \frac{\sum_{n=1}^{N} w_{l, n} h_{l, t, i, n}}{\sum_{n=1}^{N} w_{l, n}}
\end{equation}
\begin{equation}
\label{eqn:a-wavg}
\textbf{WAvg}_{l} =  LayerNorm(\textbf{A}_{l})
\end{equation}
  \item \emph{Audio-AdapterFusion (A-AF)}: given the input to the adapters at Conformer encoder layer $l$ and time step $t$, attend to different values of the hidden dimension of different task adapter outputs. A-AF combines useful knowledge from each task to solve a particular utterance.
\end{enumerate}

Below, we provide the formulas to calculate $\textbf{A-AF}_{l, t}$. For the sake of being concise, all of the following equations are for layer $l$ and time step $t$. We first project the query, key, and values to dimension $d$. We calculate the query projection $\textbf{Q}_{l, t}$ where $\textbf{H}_{l, t}$ is the stacked adapter outputs at Conformer encoder layer $l$ and time step $t$ and $\textbf{W}_l^\textbf{Q}$ is the query weight matrix at Conformer encoder layer $l$.
\begin{equation}
\label{eqn:query-projection}
\textbf{Q}_{l, t} = \textbf{H}_{l, t} \textbf{W}_l^\textbf{Q}
\end{equation}
For each task adapter $n \in \{1,...,N\}$, we calculate a different key projection $\textbf{K}_{l, t, n}$ and value projection $\textbf{V}_{l, t, n}$ where $\textbf{W}_{l, n}^\textbf{K}$ and $\textbf{W}_{l, n}^\textbf{V}$ are the key and value weight matrices, respectively, at Conformer encoder layer $l$ and task adapter $n$.
\begin{equation}
\label{eqn:key-projection}
\textbf{K}_{l, t, n} = \textbf{Z}_{l, t, n} \textbf{W}_{l, n}^\textbf{K}
\end{equation}
\begin{equation}
\label{eqn:value-projection}
\textbf{V}_{l, t, n} = \textbf{Z}_{l, t, n} \textbf{W}_{l, n}^\textbf{V}
\end{equation}
We stack the key projections and value projections of all $N$ task adapters in a new dimension.
\begin{equation}
\label{eqn:key-matrix}
\textbf{K}_{l,t} = Stack([\textbf{K}_{l, t, 0}, ..., \textbf{K}_{l, t, N}])
\end{equation}
\begin{equation}
\label{eqn:value-matrix}
\textbf{V}_{l,t} = Stack([\textbf{V}_{l, t, 0}, ..., \textbf{V}_{l, t, N}])
\end{equation}
For each value of the projection dimension $d$, get the probability distribution over the task-specific adapters and multiply it with the value projection at dimension $d$:
\begin{equation}
\label{eqn:attention}
\textbf{A}_{l, t, d} = Softmax(\textbf{Q}_{l, t} \textbf{K}_{l, t, d}^T) \textbf{V}_{l, t, d}
\end{equation}
Finally, project the adapter attention matrix $\textbf{A}_{l, t, d}$ back to the adapter output dimension with $\textbf{W}_{l, t}^\textbf{O}$ and take the LayerNorm \cite{ba2016layer} of this matrix.
\begin{equation}
\label{eqn:adapter-fusion}
\textbf{A-AF}_{l, t} = LayerNorm(\textbf{A}_{l, t} \textbf{W}_{l, t}^\textbf{O})
\end{equation}

In A-AF, weight matrices $\textbf{W}_l^\textbf{Q} \in \mathbb{R}^{d_{model} \times k}$, $\textbf{W}_l^\textbf{K} \in \mathbb{R}^{d_{model} \times N \times k}$, $\textbf{W}_l^\textbf{V} \in \mathbb{R}^{d_{model} \times N \times k}$, $\textbf{W}_l^\textbf{O} \in \mathbb{R}^{k \times d_{model}}$, and $LayerNorm$ parameters are introduced in the knowledge composition stage and trained to solve multiple tasks where $d_{model}$ is the adapter output dimension and $k$ is the projection dimension. All of the above adapter aggregation methods can be efficiently computed in matrix form using Einsum \cite{daniel2018opt}.

\subsection{Multi-Task Adapters}
In our Multi-Task Adapter (MT-A) \cite{stickland2019bert} setup, we follow the exact same learning algorithm and adapter aggregation methods as mentioned above, except all of the pre-trained single-task adapter parameters $\Phi=\{\Phi_{0}, ..., \Phi_{N}\}$ are also updated alongside the fusion parameters $\Psi$ during the knowledge composition step.

\begin{equation}
\label{eqn:mta-adapter-aggregation-loss}
\{\Phi, \Psi\} \leftarrow \argmin_{\{\Phi, \Psi\}} \mathcal{L}(D; \Theta, \Phi, \Psi)
\end{equation}

Our different adapter aggregation methods combine task-specific knowledge with varying levels of detail where the more detailed methods require more parameters. By developing different adapter aggregation methods of varying complexities, we investigate the trade-off between parameter-efficiency vs. performance of such methods and how MT-A affects this trade-off. The ideal adapter aggregation method and decision to use MT-A depends on the AI practitioner's requirements and limitations.

\section{Experiments}
\label{sec:experiments}

\subsection{Experimental Setup}
We experiment with our various adapter aggregation methods \textemdash trained with or without MT-A \textemdash to explore the trade-off between parameter-efficiency vs. ASR performance and how training in an MT-A setting affects this trade-off. We measure parameter-efficiency by the number of trained parameters during full fine-tuning or the knowledge composition stage. To investigate our models' abilities to overcome catastrophic interference, we compare our methods to the entire model fine-tuned on multiple tasks (Full FT) and an oracle (Task ID). We define the oracle as single-task adapters trained separately on multiple tasks and \emph{evaluated only on their own task}. The oracle is equivalent to using a task ID during inference time to route to adapters trained only on the specified task. 

In all of the experiments, we initialize the model with a Closed Captioning (CC) base model. The CC base model is a 120M-parameter Conformer Encoder \cite{gulati2020conformer} and a decoder with Hybrid Autoregressive Transducer (HAT) \cite{variani2020hybrid}. The CC base model is trained with fast-emit \cite{yu2021fastemit} on 17.5k hours of semi-supervised Closed Captioning data as described in Table ~\ref{tab:trainsets}. In the knowledge extraction stage, we train single-task adapters for all tasks described in Table \ref{tab:trainsets} for a maximum of 150k steps with early stopping. In our experiments, every adapter module has a bottleneck dimension of 512. In our knowledge composition experiments, we initialize the single-task adapters with the adapter parameters $\Phi$ trained in the previous stage. Weights $w_{l, n} \forall \; l\in\{1, ..., L\}, n\in\{1, ..., N\}$ in WAvg are randomly initialized with $1.0$ to weight each adapter output equally. Weights in A-AF \{$\textbf{W}_l^\textbf{Q}$, $\textbf{W}_l^\textbf{K}$, $\textbf{W}_l^\textbf{V}$, $\textbf{W}_l^\textbf{O}$\}  are initialized with Xavier initialization. Furthermore, we empirically find that a query, key, and value projection dimension of $512$ yields the best WER out of $\{256, 128, 64, 32\}$ for both A-AF with and without training in an MT-A setting.

All experiments are trained with an inverse-decay learning rate until 32k warm-up steps, then with a decay learning rate where the decay factor is $-0.5$, the initial learning rate is $3.8e^{-6}$, and the peak learning rate is $2.5e^{-4}$. We train with a global batch size of 2048 and the Adam optimizer where $\beta_{1}=0.9$, $\beta_{2}=0.999$, and $\epsilon=1e^{-6}$. We empirically find that this learning rate works well for our adapter aggregation methods. For full fine-tuning and the knowledge composition experiments, we train for a maximum of 50k steps with early stopping on the datasets described in Table~\ref{tab:trainsets}. Note that we do not train on Speech Search since we treat it as a few-shot learning task. We evaluate all of the experiments on the datasets described in Table ~\ref{tab:testsets}.

\subsection{Tasks and Datasets}
In this subsection, we will provide a general overview of the tasks and datasets used in this work. Our datasets consist of anonymized English utterances from a variety of tasks: Closed Captioning (CC), Telephony Short, Telephony Long, and Speech Search. Closed Captioning consists of subtitles or closed captioning data from an online video sharing platform, Telephony Short consists of short-form audio collected from call centers, Telephony Long consists of long-form audio collected from call centers, Agent-Only Telephony consists of agent-only, long-form audio from a contact center, and Speech Search is self-explanatory. The length of super short utterances are <2 seconds, short utterances range from 2 to 6 seconds, and long utterances range from 14 to 39 seconds.

Each utterance in the train sets is either human-labeled or pseudo-labeled. The pseudo labels are provided by a 600M-parameter teacher model trained on English Closed Captioning data with self-supervised and supervised learning \cite{peyser2023comparison}. The teacher model consists of a non-streaming Conformer encoder \cite{gulati2020conformer} with chunk-wise attention \cite{gulati2020conformer} and Connectionist Temporal Classification (CTC) loss and decoding \cite{graves2006connectionist}. We diversify the training data via multi-style training, random down-sampling from 16kHz to 8kHz, and SpecAug \cite{park19e_specaug}. Note that we do not train on Speech Search data. We describe the train set in more detail in Table ~\ref{tab:trainsets}. We evaluate our models on \emph{10 diverse test sets} from various sources with different tasks, sampling rates, utterance lengths, and number of speakers. The test sets are summarized in Table ~\ref{tab:testsets} and use WER as the scoring metric.

\begin{table}[t]
\caption{Train set descriptions including the task ID, dataset name, total number of utterances, and mean length of utterances in seconds.}
\label{tab:trainsets}
\centering
\resizebox{0.88\linewidth}{!}{%
\begin{tabular}{l|l|c|c}
    \toprule
        Task & Trainset & Total & Avg Utt \\
        ID & & Utts & Len (s) \\ \hline

        \hline
        Closed Captioning & Closed Captioning & 3500K & 18.0 \\
        \hdashline
        & Telephony Short & 315K & 3.6 \\
        Telephony Short & Telephony Super Short & 54K & 1.5 \\
        & Pseudo Telephony Short & 5298K & 4.5 \\
        \hdashline
        & Telephony Long & 266K & 14.5 \\
        Telephony Long & Agent-Only Telephony & 97K & 28.4 \\
        & Pseudo Telephony Long & 2382K & 15.9 \\
    \bottomrule
\end{tabular}
}
\end{table}

\begin{table}[t]
\caption{Test set descriptions including the task ID, dataset name, total number of utterances, and mean length of utterances in seconds.}
\label{tab:testsets}
\centering
\resizebox{0.98\linewidth}{!}{%
\begin{tabular}{l|l|c|c}
    \toprule
        Task ID & Testset & Total & Avg Utt \\
         & & Utts & Len (s) \\ \hline
        
        \hline
        Closed Captioning & Closed Captioning & 72 & 1000.0 \\
        \hdashline
         & Telephony Short 8kHz & 3615 & 2.6 \\
         & Telephony Short 16kHz & 82 & 2.6 \\
        Telephony Short & Telephony Super Short 8kHz & 3203 & 1.8 \\
         & Telephony Super Short 16kHz & 474 & 1.8 \\
         & Telephony Short Mixed kHz & 4108 & 2.6 \\
        \hdashline
         & Telephony Long 8kHz & 2000 & 15.9 \\
        Telephony Long & Telephony Long 16kHz & 3616 & 15.9 \\
         & Agent-Only Telephony & 3371 & 38.4 \\
        \hdashline
        Speech Search & Speech Search & 3000 & 5.5 \\
    \bottomrule
\end{tabular}
}
\end{table}

\subsection{Overcoming Catastrophic Interference}
As shown in Table ~\ref{tab:wer}, it is clear that \emph{Full FT suffers from catastrophic interference} when compared to the oracle (Task ID) since using a task ID to route to a set of task-specific adapters per utterance \textemdash as opposed to full fine-tuning on all the tasks \textemdash improves the WER on almost all test sets.

We investigate our task-ID-free methods' abilities to overcome catastrophic interference by comparing them to the above mentioned experiments: Full FT and Task ID. All of our proposed methods, except Avg, significantly outperform Full FT on majority or all of the test sets. Overall, \emph{A-AF, MT-A Avg, MT-A WAvg, and MT-A A-AF yield the best mean performance across all tasks with an 8\% improvement in mean WER relative to Full FT}. Furthermore, \emph{our best-performing proposed methods are on-par with using a task ID and are within 1\% mean WER relative to Task ID}. Out of these methods, \emph{A-AF slightly outperforms the others with 0\% mean WER relative to Task ID}.

Across all test sets, A-AF, MT-A Avg, MT-A WAvg, and MT-A A-AF are within 10\% WER relative to Task ID, except on the test sets Agent-Only Telephony which has a relative WER of >10\% and Telephony Short 8 kHz which has a relative WER of <-10\%. The performance of these methods on Agent-Only Telephony can potentially be explained by the lack of instances in this train set \textemdash making up only 0.5\% of the total number of training utterances. As suggested by the AdapterFusion paper \cite{pfeiffer2020adapterfusion}, we could potentially improve performance by training a separate task-specific adapter for low-resource datasets. However, the WER improvement of these methods on Telephony Short 8 kHz relative to Task ID shows that having access to adapters trained on other tasks can lead to better results on certain tasks.

To understand how our proposed models generalize to new tasks, we perform zero-shot experiments on the Speech Search testset described in Table ~\ref{tab:testsets}. As shown in Table ~\ref{tab:zero-shot-wer}, A-AF, MT-A Avg, MT-A WAvg, and MT-A A-AF outperform the base model and full fine-tuning. \emph{MT-A WAvg yields the best zero-shot performance with a 9\% WER improvement relative to full fine-tuning}.

\begin{table*}[t]
\caption{WER results of full fine-tuning and our adapter aggregation methods \textemdash with or without training in an MT-A setup \textemdash on 9 test sets from 3 diverse tasks. The test sets are separated by dashed horizontal lines to separate task IDs: Closed Captioning, Telephony Short, and Telephony Long, respectively. Each model is initialized with a 120M-parameter, CC base model. For full fine-tuning (Full FT), we fine-tune all of the weights of the base model. Adapter Mean (Avg), Adapter Weighted Mean (WAvg), and Audio-AdapterFusion (A-AF) are the different methods of aggregating adapter outputs. The A-AF architecture is illustrated in Figure ~\ref{fig:adapterfusion}. Multi-Task Adapters (MT-A) with various adapter aggregation methods show the results of jointly trained adapters and fusion weights.}
\label{tab:wer}
\centering
\resizebox{0.7\linewidth}{!}{%
\begin{tabular}{l|ccccccc|c}
\toprule

\textbf{Experiment} & \textbf{Full} & \textbf{Avg} & \textbf{WAvg} & \textbf{A-AF} & \textbf{MT-A} & \textbf{MT-A} & \textbf{MT-A} & \textbf{Task} \\
 & \textbf{FT} &  &  &  & \textbf{Avg} & \textbf{WAvg} & \textbf{A-AF} & \textbf{ID} \\
\hline
Trained Params (\#) & 120M & 0 & 36 & 27M & 21M & 21M & 48M & 21M \\ \hline

\hline

Closed Captioning & 14.9 & 44.0 & 15.5 & \textbf{14.6} & 15.1 & 15.2 & 14.7 & 14.0 \\

\hdashline

Telephony Short 8 kHz & 16.7 & 25.8 & 16.1 & 15.1 & 15.1 & 15.1 & \textbf{14.8} & 17.5 \\
Telephony Short 16 kHz & 14.6 & 24.0 & 13.6 & 13.1 & \textbf{13.0} & \textbf{13.0} & 13.1 & 14.2 \\
Telephony Super Short 8kHz & 17.9 & 21.3 & 17.8 & 15.3 & 15.2 & 15.1 & \textbf{15.0} & 15.7 \\
Telephony Super Short 16kHz & 23.9 & 28.9 & 24.0 & 22.6 & 22.1 & \textbf{21.9} & 22.1 & 20.6 \\
Telephony Short Mixed kHz & 20.6 & 35.4 & 19.7 & 18.9 & \textbf{18.8} & \textbf{18.8} & 19.1 & 19.9 \\

\hdashline

Telephony Long 8 kHz & 19.7 & 32.1 & 18.9 & 17.9 & 18.1 & \textbf{17.9} & 17.9 & 17.3 \\
Telephony Long 16 kHz & 16.5 & 40.2 & 16.8 & \textbf{15.6} & 16.4 & 16.1 & 15.7 & 15.3 \\
Agent-Only Telephony & 13.2 & 20.3 & 10.7 & 11.5 & \textbf{11.2} & 12.2 & 13.2 & 9.5 \\ \hline

\hline

\textbf{Mean} & 17.5 & 30.2 & 17.0 & \textbf{16.1} & \textbf{16.1} & \textbf{16.1} & 16.2 & 16.0 \\

\bottomrule
\end{tabular}%
}
\end{table*}

\begin{table*}[t]
\caption{Zero-shot performance of the base model (Base), full fine-tuning (Full FT), and our adapter aggregation methods \textemdash with or without training in an MT-A setup \textemdash on the Speech Search test set.}
\label{tab:zero-shot-wer}
\centering
\resizebox{0.6\linewidth}{!}{%
\begin{tabular}{l|c|ccccccc}
\toprule

\textbf{Experiment} & \textbf{Base} & \textbf{Full} & \textbf{Avg} & \textbf{WAvg} & \textbf{A-AF} & \textbf{MT-A} & \textbf{MT-A} & \textbf{MT-A} \\
 & & \textbf{FT} &  &  &  & \textbf{Avg} & \textbf{WAvg} & \textbf{A-AF}\\
\hline

Trained Params (\#) & 120M & 120M & 0 & 36 & 27M & 21M & 21M & 48M \\ \hline

\hline

Speech Search & 16.1 & 16.5 & 31.5 & 16.4 & 15.2 & 15.2 & \textbf{15.0} & 15.1 \\

\bottomrule
\end{tabular}%
}
\end{table*}

\subsection{Parameter-Efficiency}

\begin{figure}[htb]
\centering
\centerline{\includegraphics[width=0.9\columnwidth]{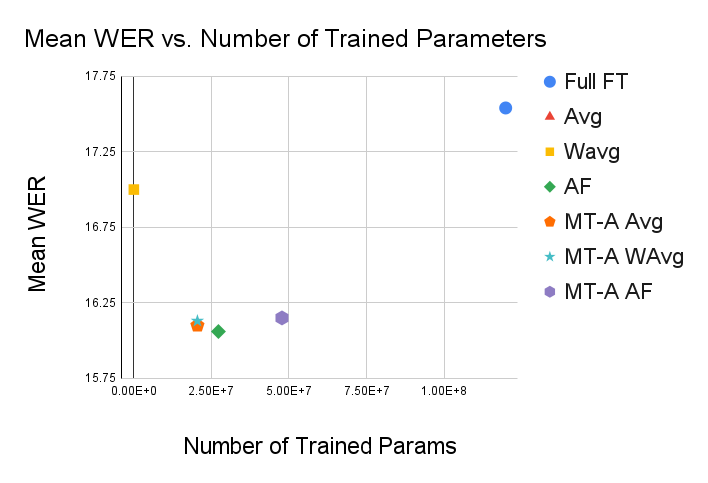}}
\caption{Parameter efficiency of different adapter aggregation methods, with and without training in an MT-A setting. Note that this plot is zoomed in for analysis and cuts off the Avg data point as the WER is above the y-axis max.}

\label{fig:parameter-efficiency}
\end{figure}
We visualize the parameter-efficiency of our proposed methods and full fine-tuning in Fig.~\ref{fig:parameter-efficiency} by plotting the mean WER over the number of trained parameters in the knowledge composition stage. Out of our proposed methods with the best performance, \emph{MT-A Avg and MT-A WAvg are the most parameter-efficient \textemdash these methods achieve within 1\% mean WER relative to Task ID, while updating only 17\% of the full model parameters in the knowledge composition stage}. However, in the scenario where pre-trained task-adapters are supplied from different vendors that prohibit the mixing of weights with other vendors or tasks, A-AF is a good alternative since A-AF freezes the adapter parameters $\Phi$ and only trains the fusion parameters $\Psi$. A-AF achieves slightly better WER performance than MT-A Avg and MT-A WAvg, but updates 23\% of the full model parameters in the knowledge composition stage. 

As expected, A-AF outperforms WAvg which outperforms Avg as seen in Table ~\ref{tab:wer}. These methods aggregate adapter outputs with progressively less detail and progressively less parameters, respectively. We notice that this is not the case for our MT-A experiments as MT-A A-AF performs slightly worse than MT-A Avg and MT-A WAvg despite training more parameters. Interestingly, the simpler adapter aggregation methods and the fewer number of trainable parameters has a regularization effect in an MT-A setting. Furthermore, we find that WAvg performs surprisingly well despite only training a 36 additional parameters in the knowledge composition stage. \emph{While only training $3e^{-7}\%$ of the full model parameters in the knowledge composition stage, WAvg achieves within 6\% mean WER relative to Task ID}. This method allows for very efficient knowledge composition and is suitable when there is limited storage space. 

\subsection{Ablations}
\begin{figure}[htb]
\centering
\centerline{\includegraphics[width=0.69\columnwidth]{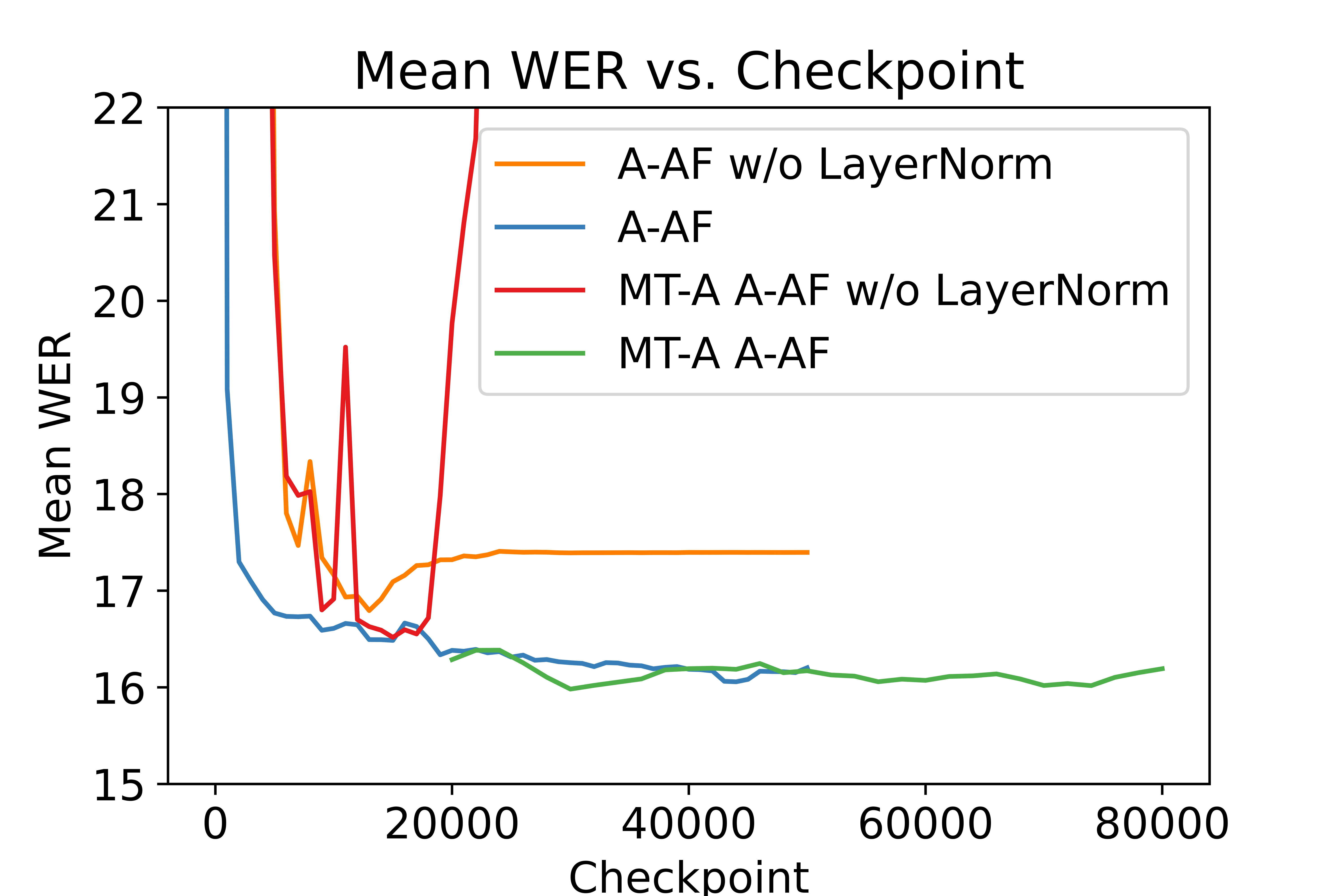}}
\caption{Mean WER vs. checkpoint for A-AF and MT-A A-AF with and without LayerNorm before the residual layer.}

\label{fig:layernorm}
\end{figure}
Since different task adapters have varying weight and output magnitudes, the magnitude of the attention output will vary depending on the probability distribution over the task-specific adapter outputs. To address this, we learn a LayerNorm to ensure that the magnitude of the signal is normalized after attending to different task adapters. As seen in Figure ~\ref{fig:layernorm}, we show that LayerNorm significantly improves performance and training stability of Audio-AdapterFusion. We yield a 4.3\% and a 3.3\% improvement in relative mean WER across all test sets in Table ~\ref{tab:wer} when LayerNorm is added before the residual layer to A-AF and MT-A A-AF, respectively.

\section{Conclusion}
\label{sec:conclusion}

In this paper, we present three novel methods to combine single-task adapters, paired with different learning algorithms, to solve multiple ASR tasks. We evaluate our methods on 10 test sets from a set of 4 diverse ASR tasks and show that our methods can significantly outperform full fine-tuning on multiple tasks and are on-par with task-ID adapter routing. We also show that our methods are parameter-efficient as they achieve comparable performance to task ID while updating only a few parameters during knowledge composition. Finally, we provide diverse combinations of adapter aggregation methods and learning algorithms to cater to different requirements and limitations of AI practitioners. In future work, we plan to explore single-task adapters trained on multiple languages to evaluate our proposed methods for multi-lingual ASR, including language mixing.

\bibliographystyle{IEEEbib}
\bibliography{refs}

\end{document}